\documentclass{article}

\usepackage{microtype}
\usepackage{graphicx}
\usepackage{subfigure}
\usepackage{booktabs} 

\usepackage{bm}
\usepackage{amsfonts}
\usepackage{epstopdf}
\usepackage{multirow}
\usepackage{mathrsfs}
\usepackage{color}
\usepackage{breakurl}

\usepackage{amsthm}
\usepackage{amssymb,amsmath}

\newcommand{\st}{\operatornamewithlimits{s.t.}}
\newcommand{\argmin}{\operatornamewithlimits{argmin}}
\newcommand{\argmax}{\operatornamewithlimits{argmax}}

\usepackage{hyperref}

\usepackage[accepted]{icml2020}

\icmltitlerunning{Loss Function Search for Face Recognition}

\begin{document}

\twocolumn[
\icmltitle{Loss Function Search for Face Recognition}



\icmlsetsymbol{equal}{*}

\begin{icmlauthorlist}
\icmlauthor{Xiaobo Wang}{equal,jd}
\icmlauthor{Shuo Wang}{equal,jd}
\icmlauthor{Cheng Chi}{casia}
\icmlauthor{Shifeng Zhang}{casia}
\icmlauthor{Tao Mei}{jd}
\end{icmlauthorlist}

\icmlaffiliation{jd}{JD AI Research}
\icmlaffiliation{casia}{Institute of Automation, Chinese Academy of Science}

\icmlcorrespondingauthor{Shifeng Zhang}{shifeng.zhang@nlpr.ia.ac.cn}

\icmlkeywords{Softmax Loss, Face Recognition}

\vskip 0.3in
]



\printAffiliationsAndNotice{\icmlEqualContribution} 

\begin{abstract}
In face recognition, designing margin-based (\textit{e.g.}, angular, additive, additive angular margins) softmax loss functions plays an important role in learning discriminative features. However, these hand-crafted heuristic methods are sub-optimal because they require much effort to explore the large design space. Recently, an AutoML for loss function search method AM-LFS has been derived, which leverages reinforcement learning to search loss functions during the training process. But its search space is complex and unstable that hindering its superiority. In this paper, we first analyze that the key to enhance the feature discrimination is actually \textbf{how to reduce the softmax probability}. We then design a unified formulation for the current margin-based softmax losses. Accordingly, we define a novel search space and develop a reward-guided search method to automatically obtain the best candidate. Experimental results on a variety of face recognition benchmarks have demonstrated the effectiveness of our method over the state-of-the-art alternatives.
\end{abstract}

\section{Introduction}
Face recognition is a fundamental and of great practice values task in the community of pattern recognition and machine learning. The task of face recognition contains two categories: face identification to classify a given face to a specific identity, and face verification to determine whether a pair of face images are of the same identity. In recent years, the advanced face recognition methods \cite{VGG,guo2018face,MV-Softmax,Arc-Softmax} are built upon convolutional neural networks (CNNs) and the learned high-level discriminative features are adopted for evaluation. To train CNNs with discriminative features, the loss function plays an important role. Generally, the CNNs are equipped with classification loss functions \cite{SphereFace,EM-Softmax,virtual,wang2019co,yao2018exploring,yao2017incorporating,guo2020learning}, metric learning loss functions \cite{Contrastive,Facenet} or both \cite{DeepID2+,Center,zheng2018ring}. Metric learning loss functions such as contrastive loss \cite{Contrastive} or triplet loss \cite{Facenet} usually suffer from high computational cost. To avoid this problem, they require well-designed sample mining strategies. So the performance is very sensitive to these strategies. Increasingly more researchers shift their attention to construct deep face recognition models by re-designing the classical classification loss functions.

Intuitively, face features are discriminative if their intra-class compactness and inter-class separability are well maximized. However, as pointed out by \cite{Center,SphereFace,AM-Softmax,Arc-Softmax}, the classical softmax loss lacks the power of feature discrimination. To address this issue, Wen \textit{et al.} \cite{Center} develop a center loss to learn centers for each identity to enhance the intra-class compactness. Wang \textit{et al.} \cite{NormFace} and Ranjan \textit{et al.} \cite{L2Constrain} propose to use a scale parameter to control the temperature of softmax loss, producing higher gradients to the well-separated samples to reduce the intra-class variance. Recently, several margin-based softmax loss functions \cite{SphereFace,virtual,cosface,AM-Softmax,Arc-Softmax} to increase the feature margin between different classes have also been proposed. Chen \textit{et al.} \cite{virtual} insert virtual classes between different classes to enlarge the inter-class margins. Liu \textit{et al.} \cite{SphereFace}  introduce an angular margin (A-Softmax) between the ground truth class and other classes to encourage larger inter-class variance. However, it is usually unstable and the optimal parameters need to be carefully adjusted for different settings. To enhance the stability of A-Softmax loss, Liang \textit{et al.} \cite{SM-Softmax} and Wang \textit{et al.} \cite{AM-Softmax,cosface} propose an additive margin (AM-Softmax) loss to stabilize the optimization. Deng \textit{et al.} \cite{Arc-Softmax}  develop an additive angular margin (Arc-Softmax) loss, which has a clear geometric interpretation. However, despite great achievements have been made, all of them are hand-crafted heuristic methods that rely on great effort from experts to explore the large design space, which is usually sub-optimal in practice.

Recently, Li \textit{et al.} \cite{AM-LFS} propose an AutoML for loss function search method (AM-LFS) from a hyper-parameter optimization perspective. Specifically, they formulate hyper-parameters of loss functions as a parameterized probability distribution sampling and achieve promising results on several different vision tasks. However, they attribute the success of margin-based softmax losses to the relative significance of intra-class distance to inter-class distance, which is not directly used to guide the design of search space. In consequence, the search space is complex and unstable, and is hard to obtain the best candidate.

To overcome the aforementioned shortcomings, including hand-crafted heuristic methods and the AutoML one AM-LFS, we try to analyze the success of margin-based softmax losses and conclude that the key to enhance the feature discrimination is to reduce the softmax probability. According to this analysis, we develop a unified formulation and define a novel search space. We also design a new reward-guided schedule to search the optimal solution. To sum up, the main contributions of this paper can be summarized as follows:

\begin{itemize}
\item{We identify that for margin-based softmax losses, the key to enhance the feature discrimination is actually \textbf{how to reduce the softmax probability}. Based on this understanding, we develop a unified formulation for the prevalent margin-based softmax losses, which involves only one parameter to be determined.}

\item{We define a simple but very effective search space, which can sufficiently guarantee the feature discrimiantion for face recognition. Accordingly, we design a random and a reward-guided method to search the best candidate. Moreover, for reward-guided one, we develop an efficient optimization framework to dynamically optimize the distribution for sampling of losses.}

\item{We conduct extensive experiments on the face recognition benchmarks, including LFW, SLLFW, CALFW, CPLFW, AgeDB, CFP, RFW, MegaFace and Trillion-Pairs, which have verified the superiority of our new approach over the baseline Softmax loss, the hand-crafted heuristic margin-based Softmax losses, and the AutoML method AM-LFS. To allow more experimental verification, our code is available at
\url{http://www.cbsr.ia.ac.cn/users/xiaobowang/}.}
\end{itemize}

\section{Preliminary Knowledge}
\noindent \textbf{Softmax}. Softmax loss is defined as the pipeline combination of last fully connected layer, softmax function and cross-entropy loss. The detailed formulation is as follows:
\begin{equation}\label{O-Softmax}
\begin{aligned}
\mathcal{L}_1 = - \log\frac{e^{\bm{w}_y^T\bm{x}}}{e^{\bm{w}_y^T\bm{x}}+\sum_{k\ne y}^Ke^{\bm{w}_k^T\bm{x}}},
\end{aligned}
\end{equation}
where $\bm{w}_k \in \mathbb{R}^d$ is the $k$-th classier ($ k \in \{1,2,\dots,K\}$) and $K$ is the number of classes. $\bm{x} \in \mathbb{R}^d$ denotes the feature belonging to the $y$-th class and $d$ is the feature dimension. In face recognition, the weights $\{\bm{w}_1,\bm{w}_2,\dots,\bm{w}_K\}$ and the feature $\bm{x}$ of the last fully connected layer are usually normalized and their magnitudes are replaced as a scale parameter $s$ \cite{NormFace,Arc-Softmax,MV-Softmax}. In consequence, given an input feature vector $\bm{x}$ with its ground truth label $y$, the original softmax loss Eq. (\ref{O-Softmax}) can be re-formulated as follows \cite{NormFace}:
\begin{equation}\label{Softmax}
\begin{aligned}
\mathcal{L}_2 = - \log\frac{e^{s\cos(\theta_{\bm{w}_y,\bm{x}})}}{e^{s\cos(\theta_{\bm{w}_y,\bm{x}})}+\sum_{k\ne y}^Ke^{s\cos(\theta_{\bm{w}_k,\bm{x}})}},
\end{aligned}
\end{equation}
where $\cos(\theta_{\bm{w}_k,\bm{x}})=\bm{w}_k^T\bm{x}$ is the cosine similarity and $\theta_{\bm{w}_k,\bm{x}}$ is the angle between $\bm{w}_k$ and $\bm{x}$. As pointed out by a great many studies \cite{L-softmax,SphereFace,AM-Softmax,Arc-Softmax,MV-Softmax}, the learned features with softmax loss are prone to be separable, rather than to be discriminative for face recognition.

\noindent \textbf{Margin-based Softmax}. To enhance the feature discrimination for face recognition, several margin-based softmax loss functions \cite{SphereFace,EM-Softmax,AM-Softmax,Arc-Softmax} have been proposed in recent years. In summary, they can be defined as follows:
\begin{equation}\label{Margin-Softmax}
\begin{aligned}
\mathcal{L}_3 = - \log\frac{e^{sf(m,\theta_{\bm{w}_y,\bm{x}})}}{e^{sf(m,\theta_{\bm{w}_y,\bm{x}})}+\sum_{k\ne y}^Ke^{s\cos(\theta_{\bm{w}_k,\bm{x}})}},
\end{aligned}
\end{equation}
where $f(m,\theta_{\bm{w}_y,\bm{x}})\leq \cos(\theta_{\bm{w}_y,\bm{x}})$ is a carefully designed margin function. Basically,
$f(m_1,\theta_{\bm{w}_y,\bm{x}})= \cos(m_1\theta_{\bm{w}_y,\bm{x}})$ is the motivation of A-Softmax loss \cite{SphereFace}, where $m_1\ge1$ and is an integer. $f(m_2,\theta_{\bm{w}_y,\bm{x}})= \cos(\theta_{\bm{w}_y,\bm{x}}+m_2)$ with $m_2>0$ is the Arc-Softmax loss \cite{Arc-Softmax}. $f(m_3,\theta_{\bm{w}_y,\bm{x}})= \cos(\theta_{\bm{w}_y,\bm{x}})-m_3$ with $m_3 >0$ is the AM-Softmax loss \cite{cosface,AM-Softmax}.  More generally, the margin function can be summarized into a combined version: $f(m,\theta_{\bm{w}_y,\bm{x}})=\cos(m_1\theta_{\bm{w}_y,\bm{x}} + m_2) - m_3$.

\noindent \textbf{AM-LFS}. Previous methods relay on hand-crafted heuristics that require much effort from experts to explore the large design space. To address this issue, Li \textit{et al.} \cite{AM-LFS} propose a new AutoML for Loss Function Search (AM-LFS) to automatically determine the search space. Specifically, the formulation of AM-LFS is written as follows:
\begin{equation}\label{AM-LFS}
\begin{aligned}
\mathcal{L}_4 = \!- \!\log \!\bigg(a_i\frac{\!e^{s\!\cos(\!\theta_{\!\bm{w}_y,\!\bm{x}})}}{\!e^{s\!\cos(\!\theta_{\!\bm{w}_y,\!\bm{x}})}+\!\sum_{\!k\ne y}^Ke^{s\!\cos(\!\theta_{\!\bm{w}_k,\!\bm{x}})}}\!+\!b_i\bigg),
\end{aligned}
\end{equation}
where $a_i$ and $b_i$ are the parameters of search space. $i \in [0, M-1]$ is the $i$-th pre-divided bin of the softmax probability. $M$ is the number of divided bins. Moreover, to consider different difficulty levels of examples, the parameters $a_i$ and $b_i$ may be different because they are randomly sampled for each bin. As a result, the search space can be viewed as a candidate set with piece-wise linear functions.

\section{Problem Formulation}
In this section, we first analyze the key to success of margin-based softmax losses from a new viewpoint and integrate them into a unified formulation. Based on this analysis, we define a novel search space and accordingly develop a random and a reward-guided loss function search method.


\subsection{Analysis of Margin-based Softmax Loss}
To begin with, let us retrospect the formulation of softmax loss Eq. (\ref{Softmax}) and margin-based softmax losses Eq. (\ref{Margin-Softmax}). The \textbf{softmax probability} $p$ is defined as follows:
\begin{equation}\label{softmaxP}
\begin{aligned}
p = \frac{e^{s\cos(\theta_{\bm{w}_y,\bm{x}})}}{e^{s\cos(\theta_{\bm{w}_y,\bm{x}})}+\sum_{k\ne y}^Ke^{s\cos(\theta_{\bm{w}_k,\bm{x}})}}.
\end{aligned}
\end{equation}
And the \textbf{margin-based softmax probability} $p_m$ is formulated as follows:
\begin{equation}\label{marginP}
\begin{aligned}
p_m = \frac{e^{sf(m,\theta_{\bm{w}_y,\bm{x}})}}{e^{sf(m,\theta_{\bm{w}_y,\bm{x}})}+\sum_{k\ne y}^Ke^{s\cos(\theta_{\bm{w}_k,\bm{x}})}}.
\end{aligned}
\end{equation}
According to the above formulations Eqs. (\ref{softmaxP}) and (\ref{marginP}), we can derive the following equation:
\begin{equation}\label{Analysis}
\begin{aligned}
p_m &= \frac{e^{sf(m,\theta_{\bm{w}_y,\bm{x}})}}{e^{sf(m,\theta_{\bm{w}_y,\bm{x}})}+\sum_{k\ne y}^Ke^{s\cos(\theta_{\bm{w}_k,\bm{x}})}} \\
& = \frac{e^{sf(m,\theta_{\bm{w}_y,\bm{x}})}}{e^{sf(m,\theta_{\bm{w}_y,\bm{x}})}+e^{\cos(\theta_{\bm{w}_y,\bm{x}})}(1-p)/p} \\
& = \frac{1}{p+e^{s[\cos({\theta}_{\bm{w}_y,\bm{x}})-f(m,\theta_{\bm{w}_y,\bm{x}})]}(1-p)}*p \\
& = \frac{1}{ap+(1-a)}*p = h(a,p)*p, \\
\end{aligned}
\end{equation}
where
\begin{equation}
a=1-e^{s[\cos({\theta}_{\bm{w}_y,\bm{x}})-f(m,\theta_{\bm{w}_y,\bm{x}})]},
\end{equation}
is a modulating factor with non-positive values ($a \leq 0$). Some existing choices are summarized in Table \ref{factors}. Particularly, when $a=0$, the margin-based softmax probability $p_m$ becomes identical to the softmax probability $p$. $h(a,p)=\frac{1}{ap+(1-a)} \in (0,1]$ is a modulating function to reduce the softmax probability. Therefore, we can claim that, no matter what kind of margin function $f(m,\theta_{\bm{w}_y, \bm{x}})$ has been designed, the key to success of margin-based softmax losses is \textbf{how to reduce the softmax probability}.


Compared to the piece-wise linear functions $p_m=a_ip+b_i$ used in AM-LFS \cite{AM-LFS}, our $p_m=h(a,p)*p$ has several advantages: 1) Our $p_m=h(a,p)*p$ is always less than the softmax probability $p$ while the piece-wise linear functions $p_m=a_ip+b_i$ are not. In other words, the discriminability of AM-LFS is not guaranteed; 2) There is only one parameter $a$ to be searched in our formulation while the AM-LFS needs search $2M$ parameters. The search space of AM-LFS is complex and unstable; 3) Our method has a reasonable range of the parameter (\textit{i.e.}, $a\leq 0$) hence facilitating the searching procedure, while the parameters of AM-LFS $a_i$ and $b_i$ are without any constraints.

\begin{table}[t]
\caption{Some existing modulating factors including Softmax, A-Softmax, AM-Softmax and Arc-Softmax, respectively.}
\label{factors}
\vskip 0.15in
\begin{center}
\begin{small}
\begin{tabular}{lcr}
\toprule
Method &  Modulating Factor $a$\\
\midrule
Softmax    & $a=0$ \\
A-Softmax     & $a=1-e^{s[\cos(\theta_{\bm{w}_y,\bm{x}})-\cos(m\theta_{\bm{w}_y,\bm{x}})]}$ \\
AM-Softmax    & $a=1-e^{sm}$ \\
Arc-Softmax   & $a=1-e^{s[\cos(\theta_{\bm{w}_y,\bm{x}})-\cos(\theta_{\bm{w}_y,\bm{x}}+m)]}$ \\
\bottomrule
\end{tabular}
\end{small}
\end{center}
\vskip -0.1in
\end{table}

\subsection{Random Search}
Based on the above analysis, we can insert a simple modulating function $h(a,p)=\frac{1}{ap+(1-a)}$ into the original softmax loss Eq. (\ref{Softmax}) to generate a unified formulation, which encourages the feature margin between different classes and has the capability of feature discrimination. In consequence, \textit{we define our search space as the choices of $h(a,p)$}, whose impacts on the training procedure are decided by the modulating factor $a$. The unified formulation is re-written as:
\begin{equation}\label{Unified}
\begin{aligned}
\mathcal{L}_5 = -\log\big(h(a,p) * p\big), \\
\end{aligned}
\end{equation}
where the modulating function $h(a,p)$ has a bounded range $(0, 1]$ and the modulating factor is $a\leq 0$. To validate our formulation Eq. (\ref{Unified}), we first randomly set the modulating factor $a \leq 0$ at each training epoch and denote this simple manner as \textbf{Random-Softmax} in this paper.

\subsection{Reward-guided Search}
The Random-Softmax can validate that the key to enhance the feature discrimination is to reduce the softmax probability. But it may not be optimal because it is without any guidance for training. To solve this problem, we propose a hyper-parameter optimization method which samples $B$ hyper-parameters $\{a_1, a_2, \dots, a_B\}$ from a distribution at each training epoch and use them to train the current model. Specifically, we model the hyper-parameter $a$ as the Gaussian distribution, described by:
\begin{equation}\label{Gaussian}
    a \sim \mathcal{N}(\mu, \sigma^2),
\end{equation}
where $\mu$ is the mean or expectation of the distribution and $\sigma$ is its standard deviation. After training for one epoch, $B$ models are generated and the rewards $\mathcal{R}(a_i), i \in [1,B]$ of these models are used to update the distribution of hyper-parameter $\mu$ by REINFORCE \cite{williams1992simple} as follows:
\begin{equation}\label{updateu}
    \mu_{e+1} = \mu_e + \eta \frac{1}{B}\sum_{i=1}^B \mathcal{R}(a_i)\nabla_{a}\log (g(a_i; \mu, \sigma)),
\end{equation}
where $g(a_i; \mu, \sigma)$ is PDF of Gaussian distribution. We update the distribution of $a$ by Eq. (\ref{updateu}) and search the best model from these $B$ candidates for the next epoch. We denote this manner as \textbf{Search-Softmax} in this paper.

\begin{algorithm}[tb]
  \caption{Search-Softmax}
  \label{Search-Softmax-a}
\begin{algorithmic}
  \STATE {\bfseries Input:} Training set $\mathcal{S}_t=\{(\bm{x}_i, y_i)\}_{i=1}^n$; Validation set $\mathcal{S}_v$; Initialized model $\mathcal{M}_{\bm{w}_0}$; Initialized distribution $\mu_0$; Distribution learning rate $\eta$; Training epochs $E$.
  \FOR{$e=1$ {\bfseries to} $E$}
  \STATE 1. Shuffle the training set $\mathcal{S}_t$  and sample $B$ hyper-parameters $\{a_1, a_2, \dots, a_B\}$ via Eq. (\ref{Gaussian});
  \STATE 2. Train the model $\mathcal{M}_{\bm{w}_e}$ for one epoch separately with the sampled hyper-parameters and get $B$ candidate models $\{\mathcal{M}_{\bm{w}_e^1}, \mathcal{M}_{\bm{w}_e^2}, \dots, \mathcal{M}_{\bm{w}_e^B}\}$;
  \STATE 3. Calculate the reward for each candidate and get the corresponding scores $\{\mathcal{R}(a_1),\mathcal{R}(a_2),\dots,\mathcal{R}(a_B)\}$;
  \STATE 4. Update the mean $\mu_{e+1}$ by using Eq. (\ref{updateu});
  \STATE 5. Decide the index of model with the highest score $i = \argmax_{j \in [1,B]} \mathcal{R}(a_j)$;
  \STATE 6. Broadcast the model $\mathcal{M}_{\bm{w}_{e+1}} = \mathcal{M}_{\bm{w}_e^i} $ ;
  \ENDFOR
  \STATE {\bfseries Output:} The model $\mathcal{M}_{\bm{w}_E}$
\end{algorithmic}
\end{algorithm}

\subsection{Optimization}
In this part, we give the training procedure of our Search-Softmax loss. Suppose we have a network model $\mathcal{M}_{\bm{w}}$ parameterized by $\bm{w}$. The training set and validation set are denoted as $\mathcal{S}_t = \{(\bm{x}_i, y_i)\}_{i=1}^n$ and $\mathcal{S}_v$, respectively. The target of our loss function search is to maximize the model $\mathcal{M}_{\bm{w}}$'s rewards $r(\mathcal{M}_{\bm{w}^*(a)}, \mathcal{S}_v)$ (\textit{e.g.}, accuracy) on the validation set $\mathcal{S}_v$ with respect to the modulating factor $a$, and the model $\mathcal{M}_{\bm{w}}$ is obtained by minimizing the following search loss:
\begin{equation}\label{Search-Softmax}
\begin{aligned}
& \max_{a} \ \mathcal{R}(a) = r(\mathcal{M}_{\bm{w}^*(a)}, \mathcal{S}_v) \\
& \st \ \  \bm{w}^*(a) = \argmin_{\bm{w}} \sum_{(\bm{x},y)\in \mathcal{S}_t} \mathcal{L}^{a}(\mathcal{M}_{\bm{w}}(\bm{x}), y).
\end{aligned}
\end{equation}
According to the works \cite{colson2007overview,AM-LFS}, the Eq. (\ref{Search-Softmax}) refers to a standard bi-level optimization problem, where the modulating factor $a$ is regarded as a hyper-parameter. We train model parameters $\bm{w}$ which minimize the training loss $\mathcal{L}^{a}$ (\textit{i.e.}, Eq. (\ref{Unified})) at the inner level, while seeking a good loss function hyper-parameter $a$ which results in a model parameter $\bm{w}^*$ that maximizes the reward on the validation set $\mathcal{S}_v$ at the outer level. The model with the highest score is used in next epoch. At last, when the training converges, we directly take the model with the highest score as the final model without any retraining. To simplify the problem, we fix $\sigma$ as a constant and optimize over $\mu$. For clarity, the whole scheme of our Search-Softmax is summarized in Algorithm \ref{Search-Softmax-a}.

\begin{table}[t]
\caption{Face datasets for training and test. (P) and (G) refer to the probe and gallery set, respectively.}
\label{data-statics}
\vskip 0.15in
\begin{center}
\begin{small}
\begin{tabular}{lccr}
\toprule
& Datasets & \#Identities   & Images     \\
\midrule
\multirow{2}{*}{Training} & CASIA-WebFace-R & 9,879 & 0.43M \\
 & MS-Celeb-1M-v1c-R & 72,690 & 3.28M	 \\
\midrule
\multirow{9}{*}{Test}
 & LFW & 5,749 & 13,233	 \\
 & SLLFW & 5,749 & 13,233	 \\
 & CALFW  & 5,749 & 12,174	 \\
 & CPLFW  & 5,749 & 11,652 \\
 & AgeDB  & 568 & 16,488	 \\
 & CFP  & 500 & 7,000 \\
 & RFW  & 11,430 & 40,607 \\
 & MegaFace  & 530 (P) & 1M (G) \\
 & Trillion-Pairs  & 5,749 (P) & 1.58M (G)\\
\bottomrule
\end{tabular}
\end{small}
\end{center}
\vskip -0.1in
\end{table}

\section{Experiments}
\subsection{Datasets}
\noindent \textbf{Training Data}. This paper involves two popular training datasets, including CASIA-WebFace \cite{Yi} and MS-Celeb-1M \cite{Msceleb}. Unfortunately, the original CASIA-WebFace and MS-Celeb-1M datasets consist of a great many face images with noisy labels. To be fair, here we use the clean version of CASIA-WebFace \cite{zhao2019multi,zhao2018towards} and MS-Celeb-1M-v1c \cite{deepglint} for training.

\noindent \textbf{Test Data}. We use nine popular face recognition benchmarks, including LFW \cite{LFW}, SLLFW\cite{SLLFW}, CALFW \cite{CALFW}, CPLFW \cite{CPLFW}, AgeDB \cite{AgeDB}, CFP \cite{CFP}, RFW \cite{RFW}, MegaFace \cite{megaface_1,megaface_2} and Trillion-Pairs \cite{deepglint}, as the test data. For more details about these test datasets, please see their references.

\noindent \textbf{Dataset Overlap Removal}. In face recognition, it is very important to perform open-set evaluation, \textit{i.e.}, there should be no overlapping identities between training set and test set. To this end, we need to carefully remove the overlapped identities between the employed training datasets and the test datasets. For the overlap identities removal tool, we use the publicly available script provided by \cite{AM-Softmax} to check whether if two names (one of which is from training set and the other comes from test set)  are of the same person. In consequence, we remove 696 identities from the training set CASIA-WebFace and 14,718 identities from MS-Celeb-1M-v1c. For clarity, we denote the refined training datasets as CASIA-WebFace-R and MS-Celeb-1M-v1c-R, respectively. Important statistics of all the involved datasets are summarized in Table \ref{data-statics}. To be rigorous, all the experiments are based on the refined training sets.

\subsection{Experimental Settings}
\noindent \textbf{Data Processing}.
We detect the faces by adopting the FaceBoxes detector \cite{facebox,zhang2019faceboxes} and localize five landmarks (two eyes, nose tip and two mouth corners) through a simple 6-layer CNN \cite{feng2017wing,liu2019high}. The detected faces are cropped and resized to 144$\times$144, and each pixel (ranged between [0,255]) in RGB images is normalized by subtracting 127.5 and divided by 128. For all the training faces, they are horizontally flipped with probability 0.5 for data augmentation.

\noindent \textbf{CNN Architecture}. In face recognition, there are many kinds of network architectures \cite{SphereFace,AM-Softmax,wang2018devil,Arc-Softmax}. To be fair, the CNN architecture should be same to test different loss functions. To to achieve a good balance between computation and accuracy, we use the SEResNet50-IR \cite{Arc-Softmax} as the backbone, which is also publicly available at the website\footnote{  \url{https://github.com/wujiyang/Face_Pytorch}}. The output of SEResNet50-IR gets a 512-dimension feature.

\noindent \textbf{Training}. Since our Search-Softmax loss is a bi-level optimization problem, our implementation settings can be divided into inner level and outer level. In the inner level, the model parameter $\bm{w}$ is optimized by stochastic gradient descent (SGD) algorithm and is trained from scratch. The total batch size is 128. The weight decay is set to 0.0005 and the momentum is 0.9. The learning rate is initially 0.1. For the CASIA-WebFace-R, we empirically divide the learning rate by 10 at 9, 18, 26 epochs and finish the training process at 30 epoch. For the MS-Celeb-1M-v1c-R, we divide the learning rate by 10 at 4, 8, 10 epochs, and finish the training process at 12 epoch. For all the compared methods, we run their source codes and keep the same experimental settings. In the outer level, we optimize the modulating factor $a$ by REINFORCE  \cite{williams1992simple} with rewards (\textit{i.e.}, accuracy on LFW) from a fixed number of sampled models. We normalized the rewards returned by each sample to zero mean and unit variance, which is set as the reward of each sample. We use Adam optimizer with a learning rate of $\eta=0.05$ and set $\sigma = 0.2$ for updating the distribution parameter $\mu$. After that, we broadcast the model parameter $\bm{w}^*$ with the highest reward for synchronization. All experiments in this paper are implemented by Pytorch \cite{paszke2019pytorch}.

\noindent \textbf{Test}. At test stage, only the original image features are employed to compose the face representations. All the reported results in this paper are evaluated by a single model, without model ensemble or other fusion strategies.

For the evaluation metric, the cosine similarity is utilized. We follow the unrestricted with labelled outside data protocol \cite{LFW} to report the performance on LFW, SLLFW, CALFW, CPLFW, AgeDB, CFP and RFW. On Megaface and Trillion-Pairs Challenge, face identification and verification are conducted by ranking and thresholding the similarity scores. Specifically, for face identification, the Cumulative Match Characteristics (CMC) curves are adopted to evaluate the Rank-1 accuracy. For face verification, the Receiver Operating Characteristic (ROC) curves are adopted. The true positive rate (TPR) at low false acceptance rate (FAR) is emphasized since in real applications false acceptance gives higher risks than false rejection.

For the compared methods, we compare our method with the baseline Softmax loss (\textbf{Softmax}) and the hand-crafted heuristic methods (including \textbf{A-Softmax} \cite{SphereFace}, \textbf{V-Softmax} \cite{virtual},  \textbf{AM-Softmax} \cite{cosface,AM-Softmax} and \textbf{Arc-Softmax} \cite{Arc-Softmax}) and one AutoML for loss function search method (\textbf{AM-LFS} \cite{AM-LFS}). For all the hand-crafted heuristic competitors, their source codes can be downloaded from the github or authors' webpages. While for AM-LFS, we try our best to re-implement it since its source code is not publicly available yet. The corresponding parameter settings of each competitor are mainly determined according to their paper's suggestions. Specifically, for V-Softmax, the number of virtual classes is set as the batch size. For A-Softmax, the margin parameter is set as $m_1=3$. While for Arc-Softmax and AM-Softmax, the margin parameters are set as $m_2=0.5$ and $m_3=0.35$, respectively. The scale parameter $s$ has already been discussed sufficiently in previous works \cite{AM-Softmax,cosface,zhang2019adacos}. In this paper, we empirically fixed it to 32 for all the methods.

\begin{figure}[ht]
\vskip 0.2in
\begin{center}
\includegraphics[width=0.494\columnwidth]{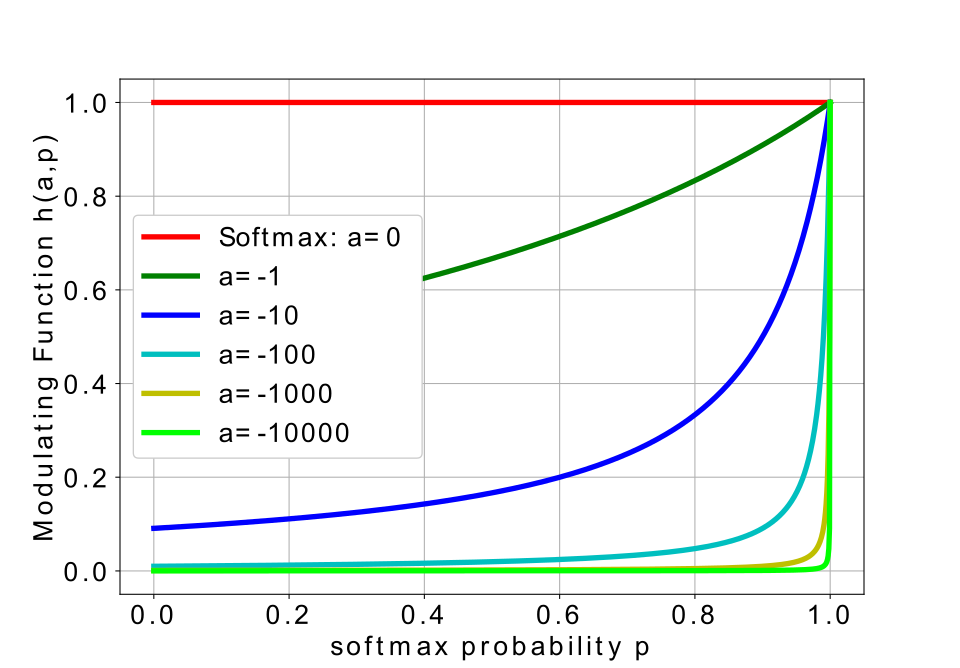}
\includegraphics[width=0.494\columnwidth]{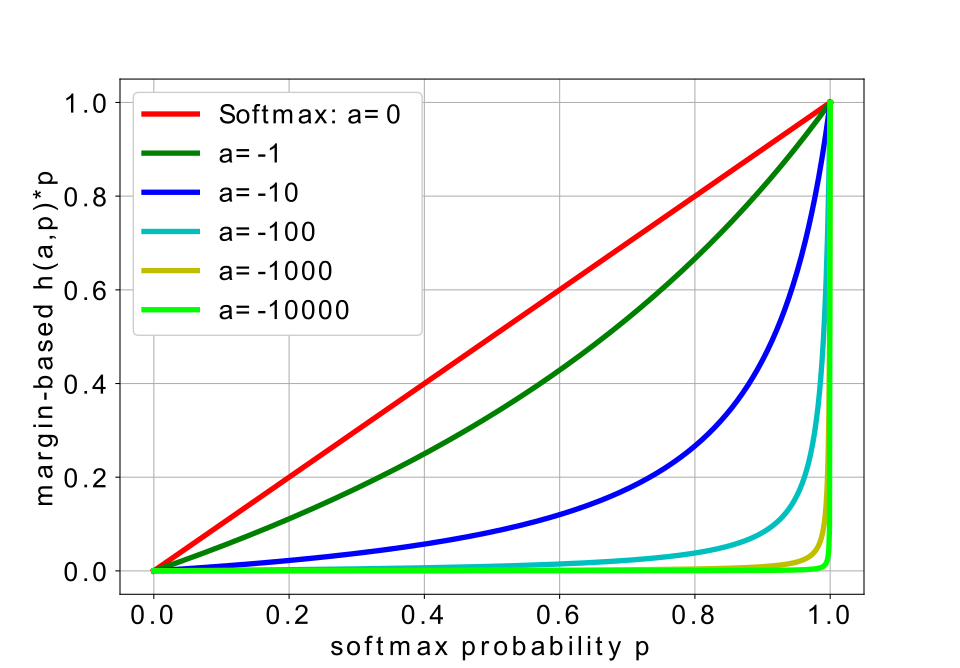}
\caption{\textbf{From Left to Right}: The derived modulating function $h(a,p)=\frac{1}{ap+(1-a)}$ and the corresponding margin-based softmax probability $p_m=h(a,p)*p$ with different modulating factors $a$. }
\label{Reduce}
\end{center}
\vskip -0.2in
\end{figure}

\subsection{Ablation Study}
\noindent{\textbf{Effect of reducing softmax probability}}. We study the effect of reducing softmax probability by setting different modulating factors $a\leq 0$. Specifically, we manually sample several values $a = \{0, -1, -10, -100, -1000, -10000\}$. The corresponding modulating functions $h(a,p)$ are shown in the left sub-figure of Figure \ref{Reduce}. From the curves, we can see that $h(a,p)$ is a monotonically increasing function with the defined domain $a \leq 0$ (\textit{i.e.}, the value of $h(a,p)$ decreases as the value of $a$ decreases). The function $h(a,p)$ is in the range $(0,1]$ hence makes $p_m=h(a,p)*p$ always less than the softmax probability $p$. The corresponding margin-based softmax probabilities $p_m=h(a,p)*p$ are displayed in the right sub-figure of Figure \ref{factor-effect}. Moreover, we also report the performance on LFW and SLLFW in Table \ref{factor-effect}. From the values, it can be concluded that reducing the softmax probability (\textit{i.e.}, $a < 0$) achieves better performance than the softmax probability (\textit{i.e.}, $a=0$). Eventually, the experiments have indicated that the key to enhance the feature discrimination is to reduce the softmax probability and give us a cue to design the search space for our Search-Softmax.

\begin{table}[t]
\caption{Effect of reducing softmax probability by setting the modulating factor $a\leq 0$. The training set is \textbf{MS-Celeb-1M-v1c-R}.}
\label{factor-effect}
\vskip 0.15in
\begin{center}
\begin{small}
\begin{tabular}{lcccccr}
\toprule
 & 0 & -1 & -10 & -100 & -1000 & -10000 \\
\midrule
LFW   & 99.53 & 99.56 & 99.66 &  99.71 & 99.61 & 99.71  \\
SLLFW & 98.78 & 98.91 & 99.20 &  99.28 & 99.36 & 99.36 \\
\bottomrule
\end{tabular}
\end{small}
\end{center}
\vskip -0.1in
\end{table}

\begin{table}[t]
\caption{Effect of the number of sampled models by setting $B$. The training set is \textbf{MS-Celeb-1M-v1c-R}.}
\label{sample-models}
\vskip 0.15in
\begin{center}
\begin{small}
\begin{tabular}{lccccr}
\toprule
 & $B=2$ & $B=4$ & $B=8$ & $B=16$ \\
\midrule
LFW    & 99.79 & 99.78 & 99.79 & 99.78 \\
SLLFW  & 99.31 & 99.56 & 99.53 & 99.58 \\
\bottomrule
\end{tabular}
\end{small}
\end{center}
\vskip -0.1in
\end{table}

\noindent{\textbf{Effect of the number of sampled models}}. We investigate the effect of the number of sampled models in the optimization procedure by changing the parameter $B$ in our Search-Softmax loss. Note that it costs more computation resources (GPUs) as $B$ increases. We report the performance results of different $B$ values selected from $\{2, 4, 8, 16\}$ in Table \ref{sample-models} in terms of accuracy on the LFW and SLLFW test sets. The results show that when $B$ is small (\textit{e.g.}, $B=2$), the performance is not satisfactory because the best candidate cannot be obtained without enough samples. We also observe that the performance exhibits saturation when we keep enlarging $B$ (\textit{e.g.}, $B\geq 4$). For a trade-off between the performance and the training efficiency, we choose to fix $B$ as 4 during training. For all the datasets, each sampled model is trained with 2 P40 GPUs, so a total of 8 GPUs are used.

\begin{figure}[ht]
\vskip 0.2in
\begin{center}
\includegraphics[width=0.85\columnwidth]{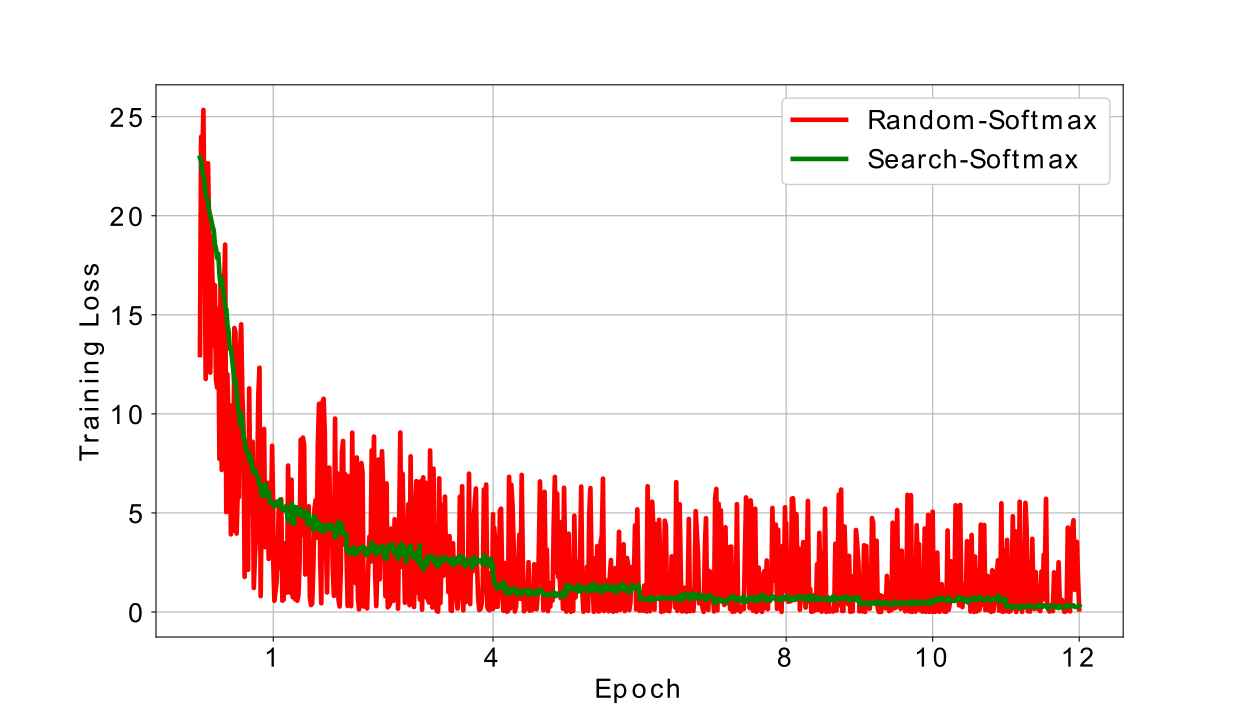}
\caption{Convergence of the proposed Random-Softmax and Search-Softmax losses. From the curves, we can see that our methods have a good behavior of convergence.}
\label{convergence}
\end{center}
\vskip -0.2in
\end{figure}

\noindent{\textbf{Convergence}}. Although the convergence of our method is not easy to be theoretically analyzed, it would be intuitive to see its empirical behavior. Here, we give the loss changes as the number of epochs increases. From the curves in Figure \ref{convergence}, it can be observed that the loss changes of our Random-Softmax is fluctuated because the modulating factor $a\leq 0$ is randomly selected at each epoch. Nevertheless, the overall trend is converged. For our Search-Softmax, we can see that it has a good behavior of convergence. The loss values obviously decrease as the  number of epochs increases and the curve is much more smooth than the Random-Softmax. The reason behind this is that our Search-Softmax updates the distribution parameter $\mu$ by the rewards of the sampled models. The parameter $\mu$ is towards optimal distribution thus the sampled $a$ for each epoch is towards decreasing the loss values to achieve better performance.

\begin{table*}[t]
\caption{Verification performance (\%) of different methods on the test sets LFW, SLLFW, CALFW, CPLFW, AgeDB and CFP. The training set is \textbf{CASIA-WebFace-R}.}
\label{CASIA-LFW}
\vskip 0.15in
\begin{center}
\begin{small}
\begin{tabular}{lccccccr}
\toprule
Method & LFW & SLLFW & CALFW & CPLFW & AgeDB & CFP & Avg. \\
\midrule
Softmax     & 97.85 & 92.98 & 86.05 & 78.58 & 89.91 & 90.58 & 89.32\\
A-Softmax   \cite{SphereFace}  & 98.30 & 93.40 & 86.36 & 78.13 & 89.43 & 90.11 & 89.28 \\
V-Softmax   \cite{virtual}     & 98.60 & 93.11 & 85.36 & 78.10 & 88.86 & 91.08 & 89.18 \\
AM-Softmax  \cite{AM-Softmax}  & 99.23 & 97.01 & 90.38 & 82.65 & 93.65 & 93.11 & 92.67 \\
Arc-Softmax \cite{Arc-Softmax} & 99.00 & 96.29 & 89.93 & 81.66 & 93.70 & 92.88 & 92.24 \\
AM-LFS \cite{AM-LFS}        & 98.88 & 95.23 & 88.14 & 80.63 & 91.41 & 92.67 & 91.16 \\
\hline
Random-Softmax (Ours) & \textbf{99.26} & 97.03 & 90.71 & 83.38 & 93.88 & 93.32 & 92.93 \\
Search-Softmax (Ours) & 99.15 & \textbf{97.68} & \textbf{90.98} & \textbf{84.21} & \textbf{94.15} & \textbf{94.21} & \textbf{93.39} \\
\bottomrule
\end{tabular}
\end{small}
\end{center}
\vskip -0.1in
\end{table*}

\begin{table*}[t]
\caption{Verification performance (\%) of different methods on the test sets LFW, SLLFW, CALFW, CPLFW, AgeDB and CFP. The training set is \textbf{MS-Celeb-1M-v1c-R}.}
\label{DG-LFW}
\vskip 0.15in
\begin{center}
\begin{small}
\begin{tabular}{lccccccr}
\toprule
Method & LFW & SLLFW & CALFW & CPLFW & AgeDB & CFP & Avg.\\
\midrule
Softmax     & 99.53 & 98.78 & 93.38 & 86.25 & 96.66 & 93.00 & 94.60 \\
A-Softmax \cite{SphereFace}    & 99.56 & 98.63 & 93.86 & 86.40 & 96.31 & 93.57 & 94.72 \\
V-Softmax \cite{virtual}       & 99.65 & 99.23 & 94.66 & 87.51 & 97.06 & 93.67 & 95.29 \\
AM-Softmax \cite{AM-Softmax}   & 99.68 & 99.40 & 95.26 & 88.63 & 97.60 & 95.22 & 95.96 \\
Arc-Softmax \cite{Arc-Softmax} & 99.69 & 99.26 & 95.21 & 88.33 & 97.35 & 95.00 & 95.80 \\
AM-LFS \cite{AM-LFS}        & 99.68 & 99.01 & 94.18 & 86.85 & 96.70 & 93.70 & 95.02 \\
\hline
Random-Softmax (Ours)  & 99.65 & 99.39 & 95.10 & 89.03 & 97.63 & 95.01 & 95.97 \\
Search-Softmax (Ours)  & \textbf{99.78} & \textbf{99.56} & \textbf{95.40} & \textbf{89.50} & \textbf{97.75} & \textbf{95.64} & \textbf{96.27} \\
\bottomrule
\end{tabular}
\end{small}
\end{center}
\vskip -0.1in
\end{table*}

\begin{table}[t]
\caption{Verification performance (\%) of different methods on the test set RFW. The training set is \textbf{CASIA-WebFace-R}.}
\label{CASIA-RFW}
\vskip 0.15in
\begin{center}
\begin{small}
\begin{tabular}{lccccr}
\toprule
Method & Caucasian & Indian & Asian & African  \\
\midrule
Softmax      & 89.16 & 77.50 & 78.16 & 74.16  \\
A-Softmax    & 88.16 & 79.33 & 79.33 & 77.50  \\
V-Softmax    & 87.00 & 78.00 & 79.49 & 73.83  \\
AM-Softmax   & 92.33 & 83.83 & 82.50 & 82.33  \\
Arc-Softmax  & 91.49 & 83.66 & 81.00 & 80.66  \\
AM-LFS       & 91.49 & 78.99 & 78.50 & 79.83  \\
\hline
Random-Softmax  & \textbf{91.99} & 84.83 & 83.66 & 82.33  \\
Search-Softmax  & 90.99 & \textbf{86.50} & \textbf{85.00} & \textbf{85.16}  \\
\bottomrule
\end{tabular}
\end{small}
\end{center}
\vskip -0.1in
\end{table}

\begin{table}[t]
\caption{Verification performance (\%) of different methods on the test set RFW. The training set is \textbf{MS-Celeb-1M-v1c-R}.}
\label{MS-RFW}
\vskip 0.15in
\begin{center}
\begin{small}
\begin{tabular}{lccccr}
\toprule
Method & Caucasian & Indian & Asian & African  \\
\midrule
Softmax      & 97.50 & 90.49 & 91.49 & 87.33  \\
A-Softmax    & 97.50 & 91.49 & 90.66 & 87.66  \\
V-Softmax    & 96.33 & 93.16 & 93.50 & 91.49  \\
Arc-Softmax  & 98.99 & 93.49 & 93.49 & 94.00  \\
AM-Softmax   & 99.00 & 95.16 & 94.66 & 94.16  \\
AM-LFS       & 91.49 & 93.49 & 92.33 & 89.99 \\
\hline
Random-Softmax  & 98.83 & 96.16 & 93.66 & 93.33  \\
Search-Softmax  & \textbf{99.00} & \textbf{96.17} & \textbf{94.67} & \textbf{95.33}  \\
\bottomrule
\end{tabular}
\end{small}
\end{center}
\vskip -0.1in
\end{table}

\subsection{Results on LFW, SLLFW, CALFW, CPLFW, AgeDB, CFP}
Tables \ref{CASIA-LFW} and \ref{DG-LFW} provide the quantitative results of the compared methods and our method on the LFW, SLLFW, CALFW, CPLFW, AgeDB and CFP sets. The bold number in each column represents the best result. For the accuracy on LFW, it is well-known that the protocol is typical and easy and almost all the competitors can achieve saturated performance. So the improvement of our Search-Softmax loss is not quite large. On the test sets SLLFW, CALFW, CPLFW, AgeDB and CFP, we can observe that our Random-Softmax loss is better than the baseline Softmax loss and is comparable to most of the margin-based softmax losses. Our Search-Softmax loss further boost the performance and is better than the state-of-the-art alternatives. Specifically, when training by the CASIA-WebFace-R dataset, our Serach-Softmax achieves about 0.72\% average improvement over the best competitor AM-Softmax. When training by the MS-Celeb-1M-v1c-R dataset, our Serach-Softmax still outperforms the best competitor AM-Softmax with 0.31\% average improvement. The main reason is that the candidates sampled from our proposed search space can well approximate the margin-based loss functions, which means their good properties can be sufficiently explored and utilized during the training phase. Meanwhile, our optimization strategy enables that the dynamic loss can guide the model training of different epochs, which helps further boost the discrimination power. Nevertheless, we can see that the improvements of our method on these test sets are not by a large margin. The reason is that the test protocol is relatively easy and the performance of all the methods on these test sets are near saturation. So there is an urgent need to test the performance of all the competitors on new test sets or test with more complicated protocols.

\begin{table}[t]
\caption{Performance (\%) of different loss functions on the test sets MegaFace and Trillion-Pairs. The training set is \textbf{CASIA-WebFace-R}.}
\label{CASIA-MegaFace}
\vskip 0.15in
\begin{center}
\begin{small}
\begin{tabular}{lccccr}
\toprule
\multirow{2}{*}{Method} & \multicolumn{2}{c}{MegaFace} &  \multicolumn{2}{c}{Trillion-Pairs} \\
& Id. & Veri. & Id. & Veri. \\
\midrule
Softmax      & 65.17 & 71.29 & 12.34 & 11.35  \\
A-Softmax    & 64.48 & 71.98 & 11.83 & 11.11 \\
V-Softmax    & 60.09 & 65.40 & 9.08 & 8.65  \\
Arc-Softmax  & 79.91 & 84.57 & 21.32 & 20.97  \\
AM-Softmax   & 82.86 & 87.33 & 25.26 & 24.66  \\
AM-LFS       & 71.30 & 77.74 & 16.16 & 15.06 \\
\hline
Random-Softmax  & 82.51 & 86.13 & 27.70 & 27.28  \\
Search-Softmax  & \textbf{84.38} & \textbf{88.34} & \textbf{29.23} & \textbf{28.49}  \\
\bottomrule
\end{tabular}
\end{small}
\end{center}
\vskip -0.1in
\end{table}

\begin{table}[t]
\caption{Performance (\%) of different loss functions on the test sets MegaFace and Trillion-Pairs. The training set is \textbf{MS-Celeb-1M-v1c-R}.}
\label{DG-MegaFace}
\vskip 0.15in
\begin{center}
\begin{small}
\begin{tabular}{lccccr}
\toprule
\multirow{2}{*}{Method} & \multicolumn{2}{c}{MegaFace} &  \multicolumn{2}{c}{Trillion-Pairs} \\
& Id. & Veri. & Id. & Veri. \\
\midrule
Softmax      & 91.10 & 92.30 & 50.34 & 46.63  \\
A-Softmax    & 90.81 & 93.49 & 49.99 & 45.59  \\
V-Softmax    & 94.45 & 95.25 & 63.85 & 61.17  \\
Arc-Softmax  & 96.39 & 96.86 & 67.60 & 66.46  \\
AM-Softmax   & 96.77 & 97.20 & 69.02 & 67.94  \\
AM-LFS       & 92.51 & 93.80 & 54.85 & 52.76 \\
\hline
Random-Softmax  & 96.15 & 96.81 & 68.73 & 68.03   \\
Search-Softmax  & \textbf{96.97} & \textbf{97.84} & \textbf{70.41} & \textbf{68.67}  \\
\bottomrule
\end{tabular}
\end{small}
\end{center}
\vskip -0.1in
\end{table}

\subsection{Results on RFW}
Firstly, we evaluate all the competitors on the recent proposed new test set RFW \cite{RFW}. RFW is a face recognition benchmark for measuring racial bias, which consists of four test subsets, namely Caucasian, Indian, Asian and African. Tables \ref{CASIA-RFW} and \ref{MS-RFW} display the performance comparison of all the involved methods. From the values, we can conclude that the results on the four subsets exhibit the same trends, \textit{i.e.}, our method is better than the baseline Softmax loss, the hand-crafted margin-based losses and the recent AM-LFS in most of cases. Concretely, our Random-Softmax obviously outperforms the Softmax loss by a large margin, which reveals that reducing the softmax probability will enhance the feature discrimination for face recognition. Our reward-guided one Search-Softmax, which defines an effective search space to well approximate the margin-based loss functions and uses rewards to explicitly search the best candidate at each epoch, is more likely to enhance the discriminative feature learning. Therefore, our Search-Softmax loss usually learns more discriminative face features and achieves higher performance than previous alternatives.

\begin{figure}[ht]
\vskip 0.2in
\begin{center}
\includegraphics[width=0.494\columnwidth]{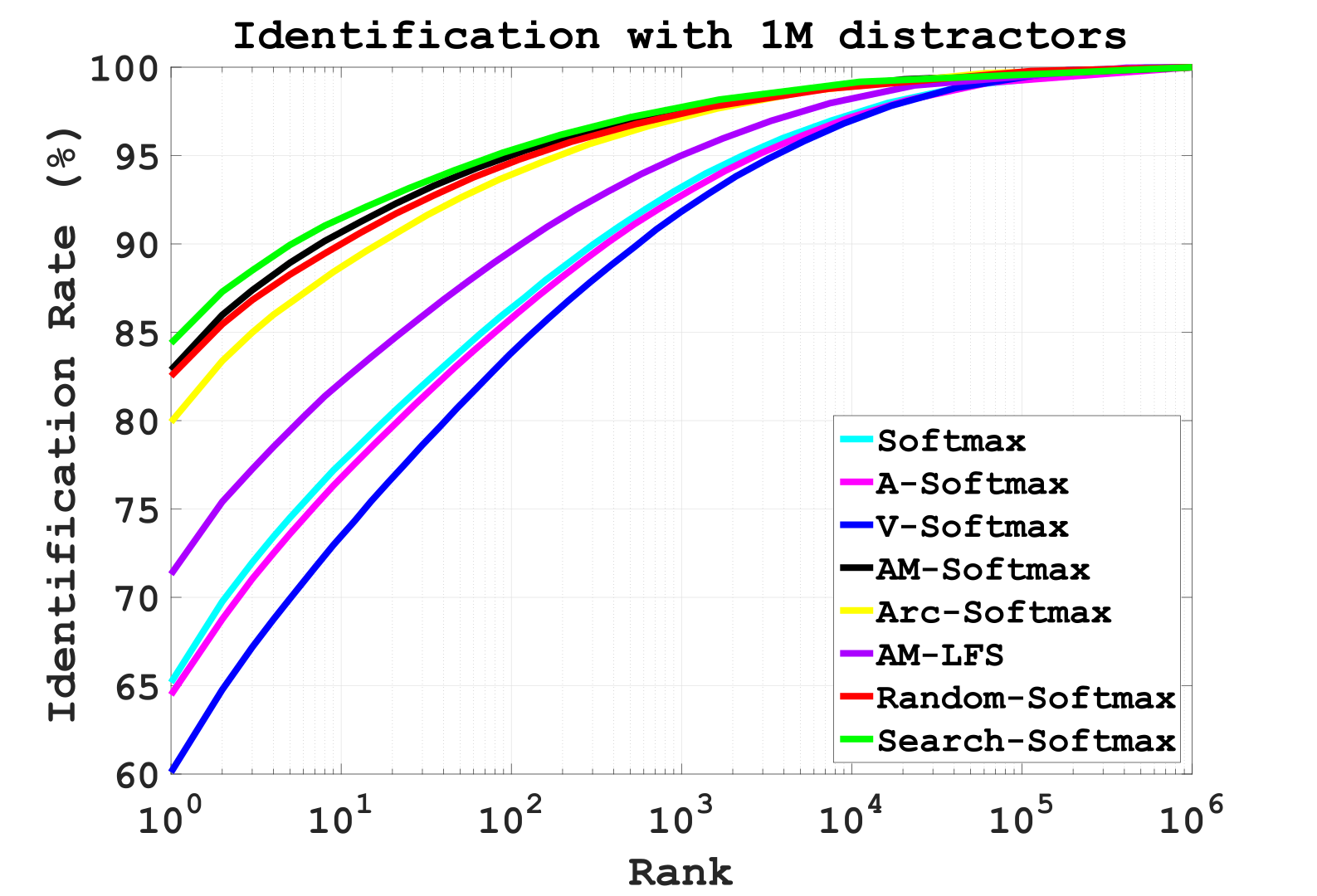}
\includegraphics[width=0.494\columnwidth]{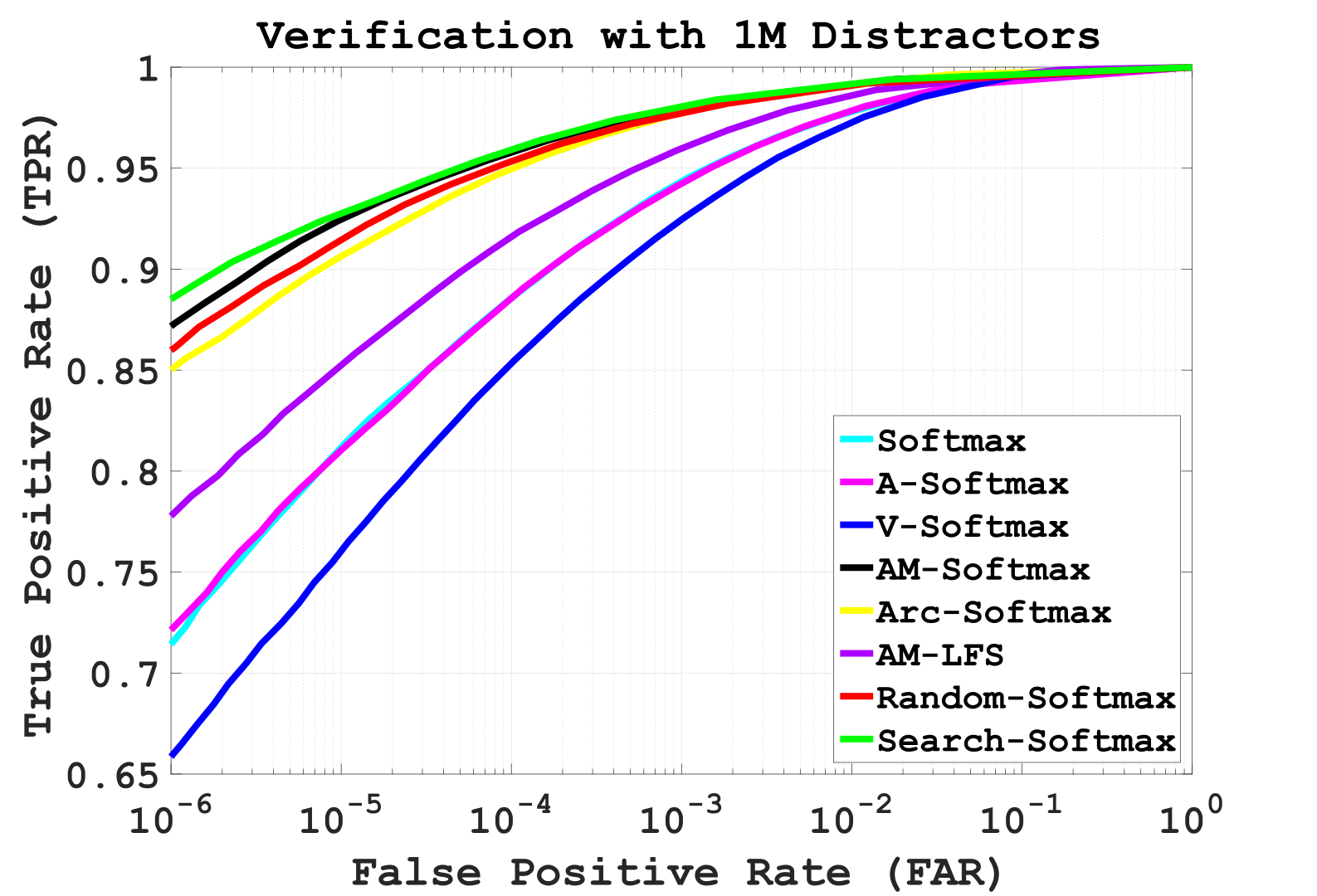}
\caption{\textbf{From Left to Right}: CMC curves and ROC curves of different loss functions with 1M distractors on MegaFace Set 1. The training set is \textbf{CASIA-WebFace-R}.}
\label{convergence-1}
\end{center}
\vskip -0.2in
\end{figure}

\begin{figure}[ht]
\vskip 0.2in
\begin{center}
\includegraphics[width=0.494\columnwidth]{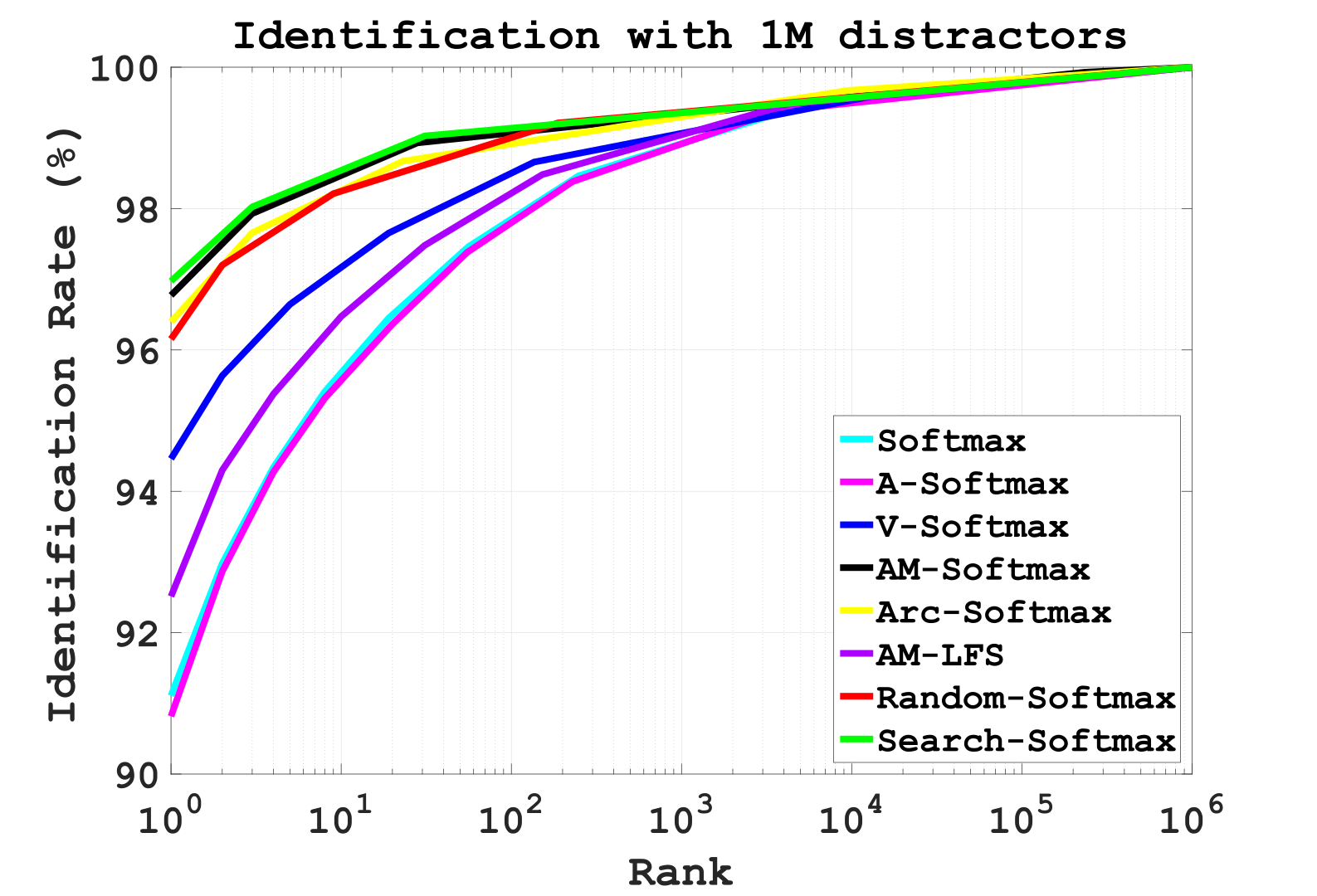}
\includegraphics[width=0.494\columnwidth]{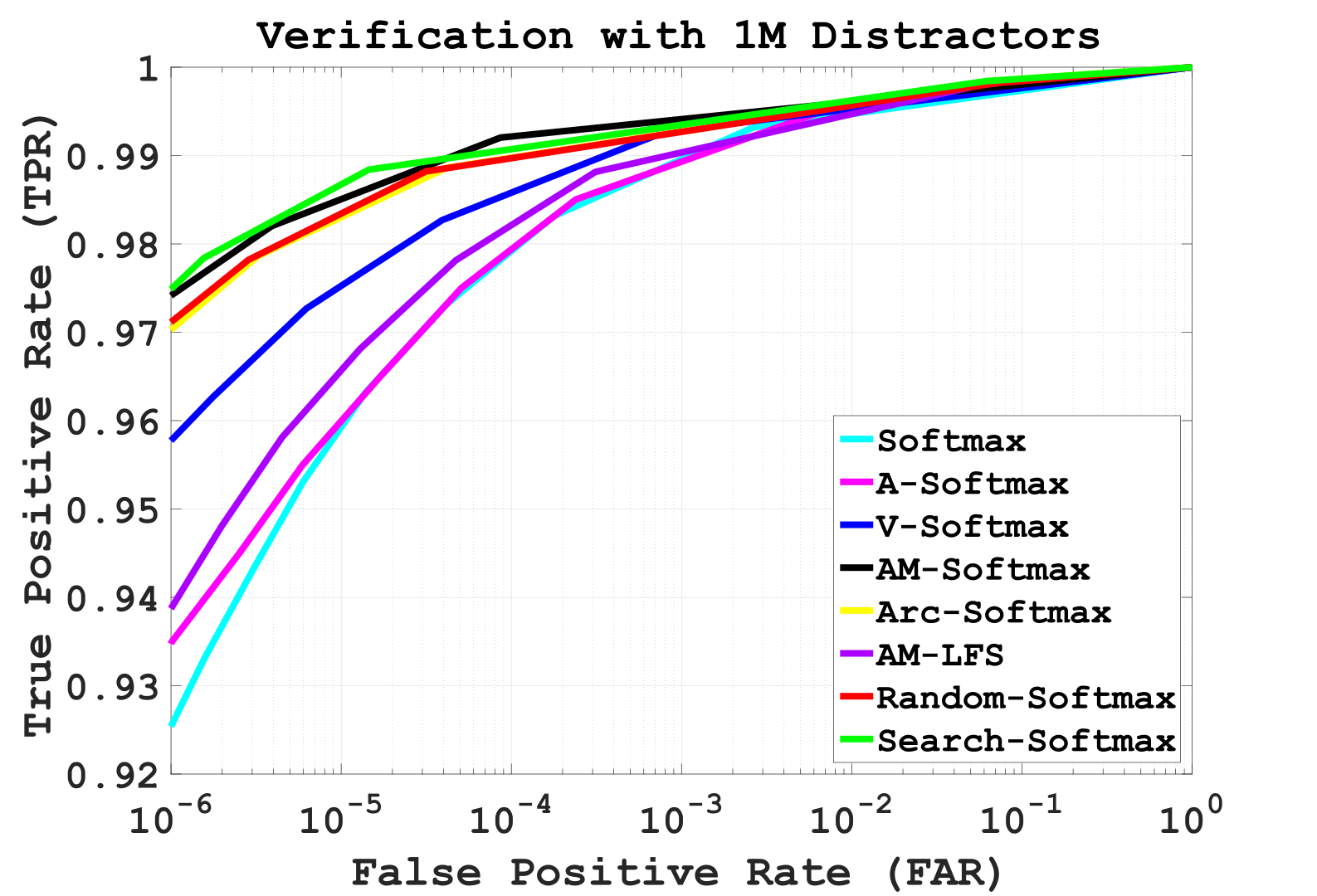}
\caption{\textbf{From Left to Right}: CMC curves and ROC curves of different loss functions with 1M distractors on MegaFace Set 1. The training set is \textbf{MS-Celeb-1M-v1c-R}.}
\label{convergence-2}
\end{center}
\vskip -0.2in
\end{figure}

\subsection{Results on MegaFace and Trillion-Pairs}
We then test all the competitors with more complicated protocols. Specifically, the identification (Id.) Rank-1 and the verification (Veri.) TPR@FAR=1e-6 on MegaFace, the identification (Id.) TPR@FAR=1e-3 and the verification (Veri.) TPR@FAR=1e-9 on Trillion-Pairs are reported in Tables \ref{CASIA-MegaFace} and \ref{DG-MegaFace}, respectively. From the numbers, we observe that our Search-Softmax achieves the best performance over the baseline Softmax loss, the margin-based softmax losses, the AutoML one AM-LFS and our naive Random-Softmax, on both MegaFace and Trillion-Pairs Challenge. Specifically, on MegaFace, for our proposed Search-Softmax, it obviously beats the best margin-based competitor AM-Softmax loss by a large margin (about 1.5\% on identification and 1.0\% on verification when training by CASIA-WebFace-R, and 0.2\% and 0.6\% when training by MS-Celeb-1M-v1c-R). Compared to the AM-LFS, our Search-Softmax loss is also better due to our new designed search space. In Figures \ref{convergence-1} and \ref{convergence-2}, we draw both of the CMC curves to evaluate the performance of face identification and the ROC curves to evaluate the performance of face verification on MegaFace Set 1. From the curves, we can see the similar trends at other measures. On Trillion-Pairs Challenge, we can observe that the results exhibit the same trends that emerged on MegaFace test set. Besides, the trends are more obvious. In particular, we achieve about 4\% improvements by CASIA-WebFace-R and 1\% improvements by MS-Celeb-1M-v1c-R at both the identification and the  verification. In these experiments, we have clearly demonstrated that our Search-Softmax loss is superior for both the identification and verification tasks, especially when the false positive rate is very low. To sum up, by designing a simple but very effective search space and using rewards to guide the discriminative learning, our new developed Search-Softmax loss has shown its strong generalization ability for face recognition.

\section{Conclusion}
This paper has summarized that the key to enhance the feature discrimination for face recognition is how to reduce the softmax probability. Based on this knowledge, we design a unified formulation for the prevalent margin-based softmax losses. Moreover, we define a new search space to guarantee the feature discrimination. Accordingly, we develop a random and a reward-guided loss function search method to obtain the best candidate. An efficient optimization framework for optimizing the distribution of search space is also proposed. Extensive experiments on a variety of face recognition benchmarks have validated the effectiveness of our new approach over the baseline softmax loss, the hand-crafted heuristic methods, \textit{i.e.}, margin-based softmax losses, and the recent AutoML one AM-LFS.

\bibliography{egbib}
\bibliographystyle{icml2020}





\end{document}